\documentclass[10pt,twocolumn,letterpaper]{article}

\usepackage{iccv}
\usepackage{times}
\usepackage{epsfig}
\usepackage{graphicx}
\usepackage{amsmath}
\usepackage{amssymb}
\usepackage{comment}
\usepackage{multirow} 
\usepackage{color}
\usepackage[breaklinks=true,bookmarks=false]{hyperref}

\iccvfinalcopy % *** Uncomment this line for the final submission

\def\iccvPaperID{****} % *** Enter the ICCV Paper ID here

\ificcvfinal\fi

\begin{document}

%%%%%%%%% TITLE
\title{Support-Set Based Cross-Supervision for Video Grounding}
% \author{\IEEEauthorblockN{AAA\IEEEauthorrefmark{2},
%   BBB\IEEEauthorrefmark{1}\IEEEauthorrefmark{3}, 
%   CCC\IEEEauthorrefmark{1}\IEEEauthorrefmark{2}\thanks{* BBB and CCC are corresponding authors.}, 
%   DDD\IEEEauthorrefmark{4}, and
%   EEE \IEEEauthorrefmark{3}}
%  \IEEEauthorblockA{\IEEEauthorrefmark{2}xxx, China, \href{mailto:xxx@xxx,xxx@xxx}{\{xxx, xxx\}@xxx}\\
%   \IEEEauthorrefmark{3}xxx, China, \href{mailto:xxx@xxx,xxx@xxx}{\{xxx, xxx\}@xxx}\\
%   \IEEEauthorrefmark{4}xxx, China,  \href{mailto:xxx@xxx}{xxx@xxx}}
% }
% \author{\IEEEauthorblockN{AAA\IEEEauthorrefmark{2},
%   BBB\IEEEauthorrefmark{1}\IEEEauthorrefmark{3}, 
%   CCC\IEEEauthorrefmark{1}\IEEEauthorrefmark{2}\thanks{* BBB and CCC are corresponding authors.}
%  }\\
%  \IEEEauthorblockA{\IEEEauthorrefmark{2}xxx, China\\
%   \IEEEauthorrefmark{3}xxx, China\\
%   \IEEEauthorrefmark{4}xxx, China}
% }
\author{Xinpeng Ding$^{1,2}$, Nannan Wang$^{1\mathrm{\ast}}$, Shiwei Zhang$^2$\thanks{Nannan Wang and Shiwei Zhang are the corresponding authors.}, De Cheng$^1$, Xiaomeng Li$^3$, \\Ziyuan Huang$^4$, 
Mingqian Tang$^2$, Xinbo Gao$^5$\\ 
{$^{1}$}Xidian University, 
{$^{2}$}Alibaba Group, 
{$^{3}$}The Hong Kong University of Science and Technology,\\
{$^{4}$}National University of Singapore,
{$^{5}$}Chongqing University of Posts and Telecommunications\\
{\tt \small xpding.xidian@gmail.com, \{nnwang,dcheng\}@xidian.edu.cn, eexmli@ust.hk}\\
{\tt \small \{zhangjin.zsw, mingqian.tmq\}@alibaba-inc.com, ziyuan.huang@u.nus.edu, gaoxb@cqupt.edu.cn}}
\maketitle
\ificcvfinal\thispagestyle{empty}\fi

%%%%%%%%% ABSTRACT
\begin{abstract}
Current approaches for video grounding propose kinds of complex architectures to capture the video-text relations, and have achieved impressive improvements.
However, it is hard to learn the complicated multi-modal relations by only architecture designing in fact.
In this paper, we introduce a novel Support-set Based Cross-Supervision (Sscs) module which can improve existing methods during training phase without extra inference cost.
The proposed Sscs module contains two main components, \emph{i.e.}, discriminative contrastive objective and generative caption objective.
The contrastive objective aims to learn effective representations by contrastive learning, while the caption objective can train a powerful video encoder supervised by texts.
Due to the co-existence of some visual entities in both ground-truth and background intervals, \emph{i.e.} mutual exclusion, naively contrastive learning is unsuitable to video grounding.
We address the problem by boosting the cross-supervision with the support-set concept, which collects visual information from the whole video and eliminates the mutual exclusion of entities.
Combined with the original objectives, Sscs can enhance the abilities of multi-modal relation modeling for existing approaches.
We extensively evaluate Sscs on three challenging datasets, and show that our method can improve current state-of-the-art methods by large margins, especially 6.35\% in terms of R1@0.5 on Charades-STA.

\end{abstract}

%%%%%%%%% BODY TEXT
\section{Introduction}

\begin{figure}
    \centering
    \includegraphics[width=0.9\columnwidth,height=0.34\textheight]{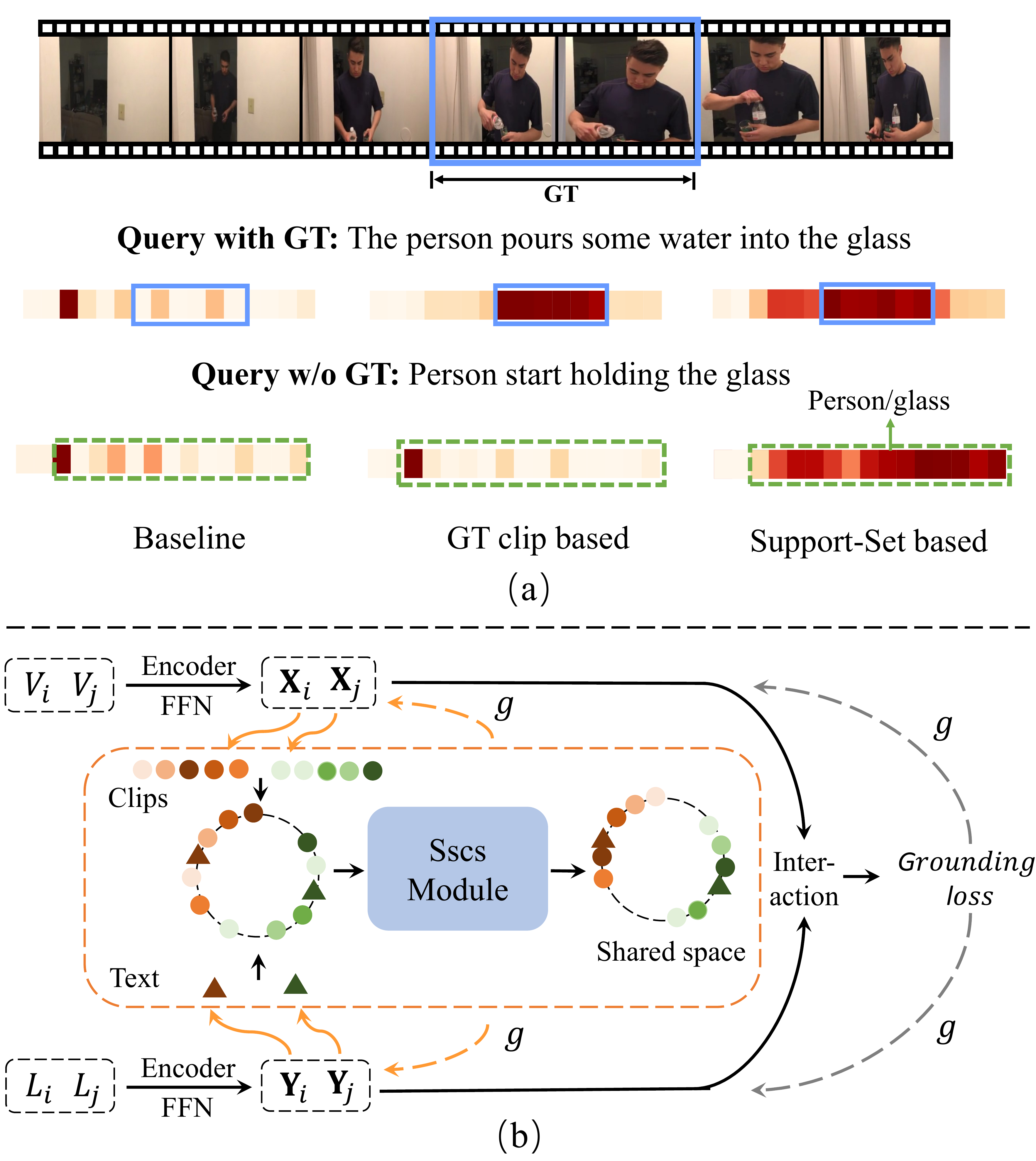}
    \caption{(a) Comparison of the attention map of the similarity between video clips and text queries.
    The darker the color, the higher the similarity.
    `GT' indicates the ground-truth.
    (b) The proposed Support-Set based Cross-Supervision (Sscs) Module. Sscs makes the embedding of semantic-related clip-text pairs (dark circles and triangles) to be close in the shared feature space. 
    }
    \label{fig:motivation}
\end{figure}

Video grounding aims to localize the target time intervals in an untrimmed video by a text query.
As illustrated in Fig.~\ref{fig:motivation} (a), given a sentence `The person pours some water into the glass.' and a paired video, the target is to localize the best matching segment, \emph{i.e.}, from 7.3s to 17.3s.
%
%-------------------------------------------------------------------------
%----------------------------------------------------------------------------
%
There are various methods~\cite{zhang2019learning,zeng2020dense,gao2017tall} have been proposed for this task, and they have made significant progresses.
These methods can reach an agreement that video-text relation modeling is one of the crucial roles.
An effective relation should be that semantically related videos and texts must have high responses, and vice versa.

To achieve this goal, existing methods focus on carefully designing complex video-text interaction modules.
For example, Zeng~\emph{et al.}~\cite{zeng2020dense} propose a pyramid neural network to consider multi-scale information.
Local-global strategy~\cite{mun2020local} and self-modal graph attention~\cite{liu2020jointly} are applied as the interaction operations to learning the multi-modal relations.
After that, they use the interacted features to perform video grounding straightway.
However, the multi-modal relations are complicated because the video and text have unequal semantics, \emph{e.g.}, `person' is just one word but may last a whole video.
Hence, existing methods based on the architecture improvements have limited capacities to learn video-caption relations; see Fig.~\ref{fig:motivation} (a) (Please see `Baseline').

\begin{figure}
    \centering
    \includegraphics[width=1.0\columnwidth,height=0.13\textheight]{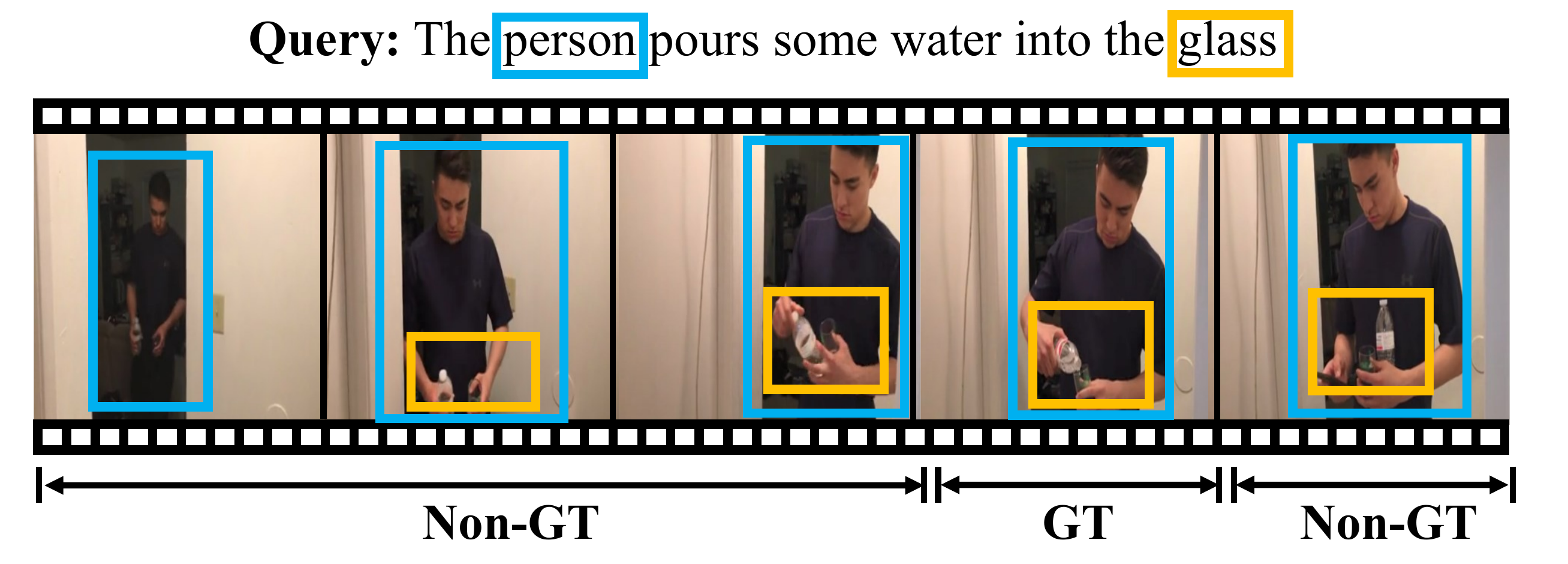}
    \caption{Mutual exclusion of entities.
    The `person' and `glass' entities appear in both ground-truth (GT) clips and non-ground-truth clips (Non-GT).
    Although there are no `pour water' action happening in Non-GT clips, the semantics of the Non-GT video clips are also similar with those of GT ones, due to the common entities.
    }
    \label{fig:mutualexclusion}
\end{figure}

Motivated by the advances of multi-modal pre-training \cite{miech2020end,patrick2020support,miech2019howto100m}, we propose a Support-Set Based Cross-Supervision, termed Sscs, to improve multi-modal relation learning for video grounding in a supervision way compared with the hand-designed architectures.
As shown in Fig.~\ref{fig:motivation}, the Sscs module is an independent branch that can be easily embedded into other approaches in the training stage.
The proposed Sscs includes two main components, \emph{i.e.}, contrastive objective and caption objective.
The contrastive objective is as typical discriminative loss function, that targets to learn multi-modal representations by applying infoNCE loss function \cite{miech2020end,patrick2020support}.
In contrast, the caption objective is a generative loss function, which can be used to train a powerful video encoder \cite{goodfellow2014generative,zhou2018end}.
For an untrimmed video, there are some vision entities appear in both ground-truth and background intervals, \emph{e.g.}, the person and glass in Fig.~\ref{fig:mutualexclusion}, but the original contrastive learning may wipe away the same parts between the foreground and background, including the vision entities.
These vision entities are also important for video grounding task, \emph{e.g.}, thus it is unsuitable to directly apply the contrastive learning into the video grounding task directly.
To solve this problem, we apply the support-set concept, which captures visual information from the the whole video, to eliminates the mutual exclusion of entities.
By this means, we can improve the cross-supervision module naturally and further enhance the relation modeling.
To prove the robustness, we choose two state-of-the-art approaches as our baselines, \emph{i.e.}, 2D-TAN~\cite{zhang2019learning} and LGI~\cite{mun2020local}, and the experimental results show that the proposed Sscs can achieve a remarkable improvement.
Our contributions are summarized as three-folds: {\bf (a)} We introduce a novel cross-supervision module for video grounding, which can enhance the correlation modeling between videos and texts but not bring in the extra inference cost. {\bf (b)} We propose to apply support-set concept to address the mutual exclusion of video entities, which make it is more suitable to apply contrastive learning for video grounding. {\bf (c)} Extensive experiments illustrate the effectiveness of Sscs on three public datasets, and the results show that our method can significantly improve the performance of the state-of-the-art approaches.

\section{Related Work}
\noindent{\bf Video grounding.} 
Early approaches \cite{gao2017tall,anne2017localizing,xu2019multilevel,ge2019mac} for video grounding use a two-stage visual-textual matching strategy to tackle this problem, which require a large number of proposals.
It is important for these methods to improve the quality of the proposals.
SCDM~\cite{yuan2019semantic} incorporates the query text into the visual feature for correlating and composing the sentence-related video contents over time. 
2D-TAN \cite{zhang2019learning} adopts a 2D temporal map to model temporal anchors, which can extract the temporal relations between video moments. 
To process more efficiently, recently, many one-stage methods \cite{zeng2020dense,liu2020jointly,yuan2019find,he2019read,wang2019language,xia2019anchor} are proposed to predict starting and ending times directly. 
Zeng {\it et al.} \cite{zeng2020dense} avoid the imbalance training by leveraging much more positive training samples, which improves the grounding performance. 
LGI \cite{mun2020local} improves the performance of localization by exploiting contextual information from local to global during bi-modal interactions.
%
%-------------------------------------------------------------------------
%----------------------------------------------------------------------------

\begin{figure*}
    \centering
    \includegraphics[width=1.9\columnwidth,height=0.23\textheight]{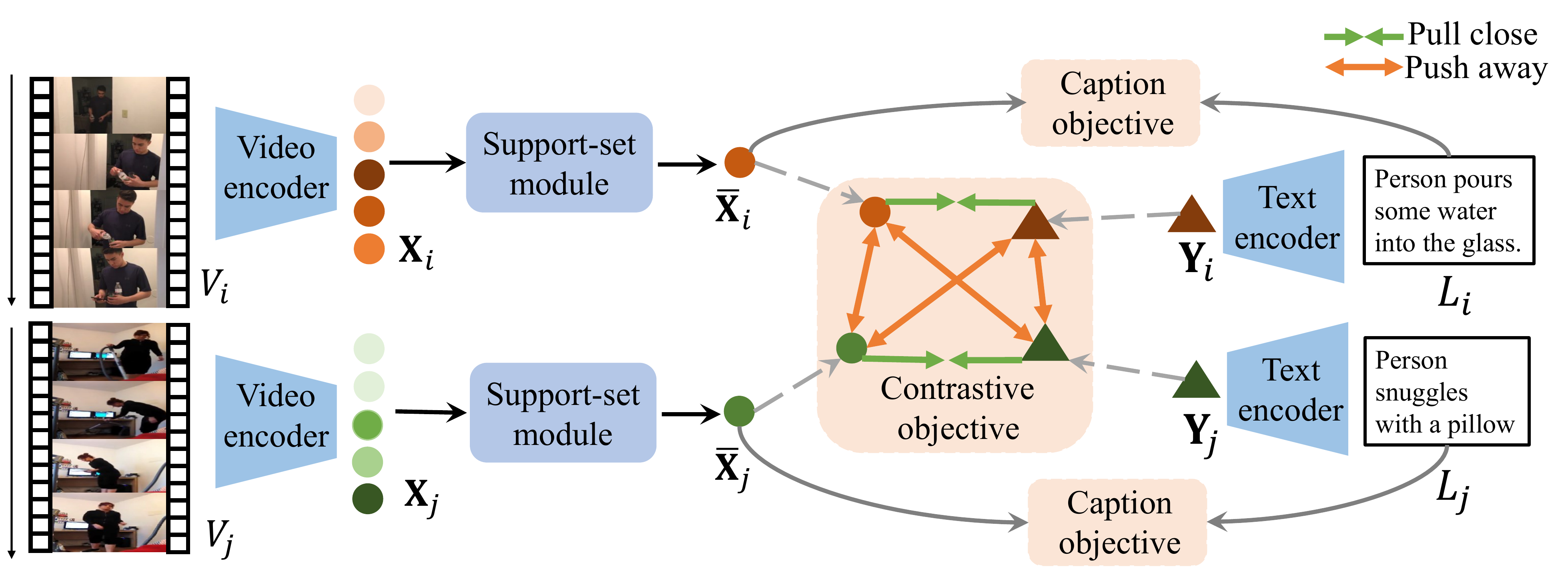}
    \caption{
    Illustration of our proposed Support-set Based Cross-Supervision Module.
    For clarity, we only present two video-text pairs $\{V_i, L_i \}$, $\{V_j, L_j \}$ in the batch.
    After feeding them into a video and text encoder, the clip-level and sentence-level embedding ($\{ \mathbf{X}_i, \mathbf{Y}_i \}$ and $\{ \mathbf{X}_j, \mathbf{Y}_j \}$) in a shared space are acquired.
    Base on the support-set module (see details in Fig.~\ref{fig:comparision} (b) ), we compute the weighted average of $\mathbf{X}_i$ and $\mathbf{X}_j$ to obtain $\overline{\mathbf{X}}_i$, $\overline{\mathbf{X}}_j$ respectively.
    Finally, we combine the contrastive and caption objectives to pull close the representations of the clips and text from the same samples and push away those from other pairs.
    }
    \label{fig:Sscs}
\end{figure*}

\noindent{\bf Multi-modal Representation Learning.} 
A mass of self-supervised methods \cite{chen2020exploring,benaim2020speednet,devlin2018bert} have been proposed to pre-train models on large-scale multi-modal data, such as images~\cite{russakovsky2015imagenet}, videos~\cite{carreira2017quo} and text~\cite{zhu2015aligning}.
To learn video-text representations, a large-scale instructional video dataset, HowTo100M \cite{miech2019howto100m}, is released.
Some works use the contrastive loss to improve video-text representations based on HowTo100M for tasks such as video caption \cite{zhou2018end}, video retrieval \cite{araujo2017large} and video question answering \cite{lei2018tvqa}.
MIL-NCE~\cite{miech2020end} brings the multi instance learning into the contrastive learning framework to address the misalignment between video content and narrations.
Patrick {\it et al.}~\cite{patrick2020support} combine both discriminative and generative objectives to push related video and text instance together.
Compared with these approaches, our method targets to improve video grounding via multi-modal training without extra inference cost.
%-------------------------------------------------------------------------

%-------------------------------------------------------------------------

%-------------------------------------------------------------------------
\section{Proposed Method}
%-------------------------------------------------------------------------

\subsection{Problem Formulation}
\label{sec:formulation}
Let's define a set of video-text pairs as $\mathcal{C}=\{(V_i,L_i)\}_{i=1}^N$, where $N$ is the number of video-text pairs, $V_i$ and $L_i$ are the $i$-th untrimmed video and sentence respectively. 
Given a query sentence $L_i$, the purpose of video grounding is to localize a target time interval $\mathcal{A}_i=(t^i_s,t^i_e)$ in $V_i$, where $t^i_s$ and $t^i_e$ denote the starting and ending time respectively. 
%

%-------------------------------------------------------------------------
\subsection{Video and Sentence Encoding}
\label{sec:embedding}
\noindent\textbf{Video encoding.} 
We first divide a long untrimmed video $V_i$ into $T$ clips, defined as $V_i=\{v^i_t\}_{t=1}^T$. 
Each clip consists of a fixed number of frames. 
Then, $T$ clips are fed into a pre-trained 3D CNN model to extract the video features $\mathbf{F}_i = \{ \mathbf{f}_t^i \} \in \mathbb{R}^{T \times D_v}$, where $D_v$ denotes the dimension of the clip-based video features.

\noindent\textbf{Sentence encoding.} 
For $m$-th word $l^i_m$ in a sentence $L_i$, we feed it into the GloVe word2vec model \cite{pennington2014glove} to obtain the corresponding word embedding $w^i_m$. 
Then, the word embeddings are sequentially fed into one three layer bidirectional LSTM network \cite{hochreiter1997long}, and we use its last hidden state as the features of the sentence $L_i$, \emph{i.e.}, $\mathbf{G}_i \in \mathbb{R}^{D_l}$.
%-------------------------------------------------------------------------
\subsection{Cross-Supervised Video Grounding}
\label{sec:ssmmobj}

In this section, we first outline the overall framework in Section \ref{sec:architecture}. 
Then, in Section \ref{sec:supportset}, we introduce the concept of support-set for video grounding in details. 
Finally, we introduce several kinds of support-set in Section \ref{sec:differentss}.

%------------------------------------------------------------------------
\subsubsection{The Overall Framework}
\label{sec:architecture}

The key of video grounding is to capture the relations between videos and texts.
That is, it should have a high similarity between $V_i$ and $S_j$ if they are semantically-related and vice versa.
For this purpose, most existing methods design multitudinous architectures to capture the relations by modeling the video-text interactions~\cite{zhang2019learning, mun2020local}.
Typically, they first fuse the visual and textual embeddings $\mathbf{X}_i$ and $\mathbf{Y}_i$, and then predict target time intervals $\hat{\mathcal{A}}=\left( \hat{t}^i_s, \hat{t}^i_e \right)$ directly.
At the training stage, the loss function $\mathcal{L}^{\text{vg}}$ is applied on the fused features to optimize the models.
The function $\mathcal{L}^{\text{vg}}$ may have different form in different method, \emph{e.g.}, binary cross entropy loss function applied in 2D-TAN~\cite{zhang2019learning}.

Unlike these methods, we introduce two cross-supervised training objectives that can improve existing methods just during training phase.
The two objectives contain a contrastive objective and a caption objective, and can be insert existing methods directly.
Thus the overall framework mainly contains two components, \emph{i.e.}, the commonly used video grounding framework and the proposed cross-supervised objectives.
Hence, the overall objective of our method is:
\begin{equation}
    \mathcal{L} = \mathcal{L}^{\text{vg}} + \lambda_1 \mathcal{L}^{\text {contrast}} + \lambda_2 \mathcal{L}^{\text {caption}},
\label{e:loss}
\end{equation}
where $\mathcal{L}^{\text {contrast }}$ and $\mathcal{L}^{\text {caption}}$ denote the contrastive objective and caption objective respectively.
The hyperparameters $\lambda_1$ and $\lambda_2$ control the weight of two objectives.

\subsubsection{Cross-Supervised Objectives}\label{sec:supportset}

The target of the cross supervised objectives is to learn effective video-text relations, as illustrated in Fig.~\ref{fig:Sscs}. 
To make it clear, we first introduce the GT clip-based learning, based on which we present the details of the proposed cross-supervised objectives.
After that, we discus existing problems, \emph{i.e.}, the mutual exclusion between the visual and textual modalities.
Finally, we provide the solution by support set based learning.

\noindent{\bf GT Clip-Based Learning.} 
In video grounding, a sentence usually corresponds to multiple clips, which are all contained in a ground-truth interval.
An intuitive method to learn a powerful representation is to set the clips in ground-truth (GT) intervals as the positive samples, while others are negatives, \emph{i.e.}, clips in Non-GT intervals and other videos.
Formally, we denote A {\it mini-batch} of samples from $\mathcal{C}$ with $\mathcal{B}$, hence $\mathcal{B} \subset \mathcal{C}$. 
Then the samples in the {\it mini-batch} can be defined as $\mathcal{B}=\{(V_i,L_i)\}_{i=1}^B$, where $B$ is the size of the {\it mini-batch}.
After fed into the video and text encoders, we can obtain base embeddings $\{(\mathbf{F}_i,\mathbf{G}_i)\}_{i=1}^B$. 
Then the embeddings can be mapped into a same space with equal dimension by $\mathbf{X}_i= \Psi\left( \mathbf{F}_i \right)$ and $\mathbf{Y}_i = \Phi \left( \mathbf{G}_i \right)$.
For a pair of the video and text embeddings $(\mathbf{X}_i,\mathbf{Y}_i)$ in $\mathcal{B}$, we define the set of ground-truth clips as $\mathcal{M}_i = \{ \mathbf{x}^i_t \, | \, t \in [t^i_s, t^i_e] \}$, where $t^i_s$ and $t^i_e$ denote the starting and ending time of the ground-truth, $\mathbf{x}^i_t $ is the $t$-th vector in $\mathbf{X}_i$.
The set of background clips for $V_i$ can be denoted as $\overline{\mathcal{M}}_i = \{ \mathbf{x}^i_t \, | \, t \notin [t^i_s, t^i_e] \}$.
Then, the positive pairs in $\mathcal{B}$ can be constructed by considering the ground-truth clips together with the corresponding text, defined as $\mathcal{P}_i = \{  (\mathbf{x}, \mathbf{Y}_i) \, | \, \mathbf{x} \in \mathcal{M}_i  \}$.
The non GT clips and clips in other videos can be regarded as the negative samples of the text $L_i$, i.e, $\mathcal{N}_i = \{  (\mathbf{x}, \mathbf{Y}_i) \, | \, \mathbf{x} \in \overline{\mathcal{M}}_i \cup \mathbf{X}_j, \, i \neq j \}$.

\noindent{\bf Contrastive objective.}
Based on the above definitions, we detail the contrastive objective here.
The purpose of the contrastive objective is to learn effective video-text representations, for which we use a contrastive loss to increase the similarities of positive pairs in $\mathcal{P}$ and push away those from the negative pairs in $\mathcal{N}$.
Specifically, we minimize the softmax version of MIL-NCE~\cite{miech2020end} as follows:

\begin{equation}
      -\sum\limits_{i=1}^{B} \log \left(\frac{ \sum\limits_{\left(\mathbf{x}, \mathbf{y} \right) \in \mathcal{P}_{i}} e^{ \mathbf{x}^{\top} \mathbf{y} / \tau} }{\sum\limits_{\left(\mathbf{x}, \mathbf{y} \right) \in \mathcal{P}_{i}}   e^{\mathbf{x}^{\top} \mathbf{y} / \tau } +\sum\limits_{\left(\mathbf{x}^{\prime}, \mathbf{y}^{\prime}\right) \in \mathcal{N}_{i}} e^{ {\mathbf{x}^{\prime}}^{\top} {\mathbf{y}^{\prime}} / \tau}  }\right),
     \label{e:milnce}
\end{equation}

where $\tau$ is the temperature weight to control the concentration level of the sample distribution \cite{hinton2015distilling}. 
Therefore, the constrastive objective is a typical kind of discriminative loss function.

\noindent{\bf Caption objective.}
Besides the contrastive objective, we also introduce the caption objective~\cite{patrick2020support,korbar2020video} to further improve the video-text representation.
The caption objective can be formulized as:

\begin{equation}
    \mathcal{L}^{\text {caption}} = -\frac{1}{B} \sum_{i=1}^{B} \log p\left(l_{i} \mid \mathbf{w}_{i}\right),
    \label{e:caption}
\end{equation}

where $l_i$ is the $i$-th word of $L_i$, $\mathbf{w}_{i} \in \mathbb{R}^D$ is the embedding for generating the sentence, which is obtained by $\mathbf{w}_i = \Phi^{\prime}\left(\mathbf{X}^{gt}_{i} \right)$.
$\mathbf{X}^{gt}_i$ is the concatenated features in ground-truth clips, \emph{i.e.}, $\mathbf{X}^{gt}_i=[\mathbf{x}^i_{t^i_s},..,\mathbf{x}^i_{t^i_e}]$. $\Phi^{\prime}$ is the transformation layer which can be convolutional layers \cite{simonyan2014very} or self-attention \cite{vaswani2017attention}.

\begin{figure}
    \centering
    \includegraphics[width=1\columnwidth,height=0.25\textheight]{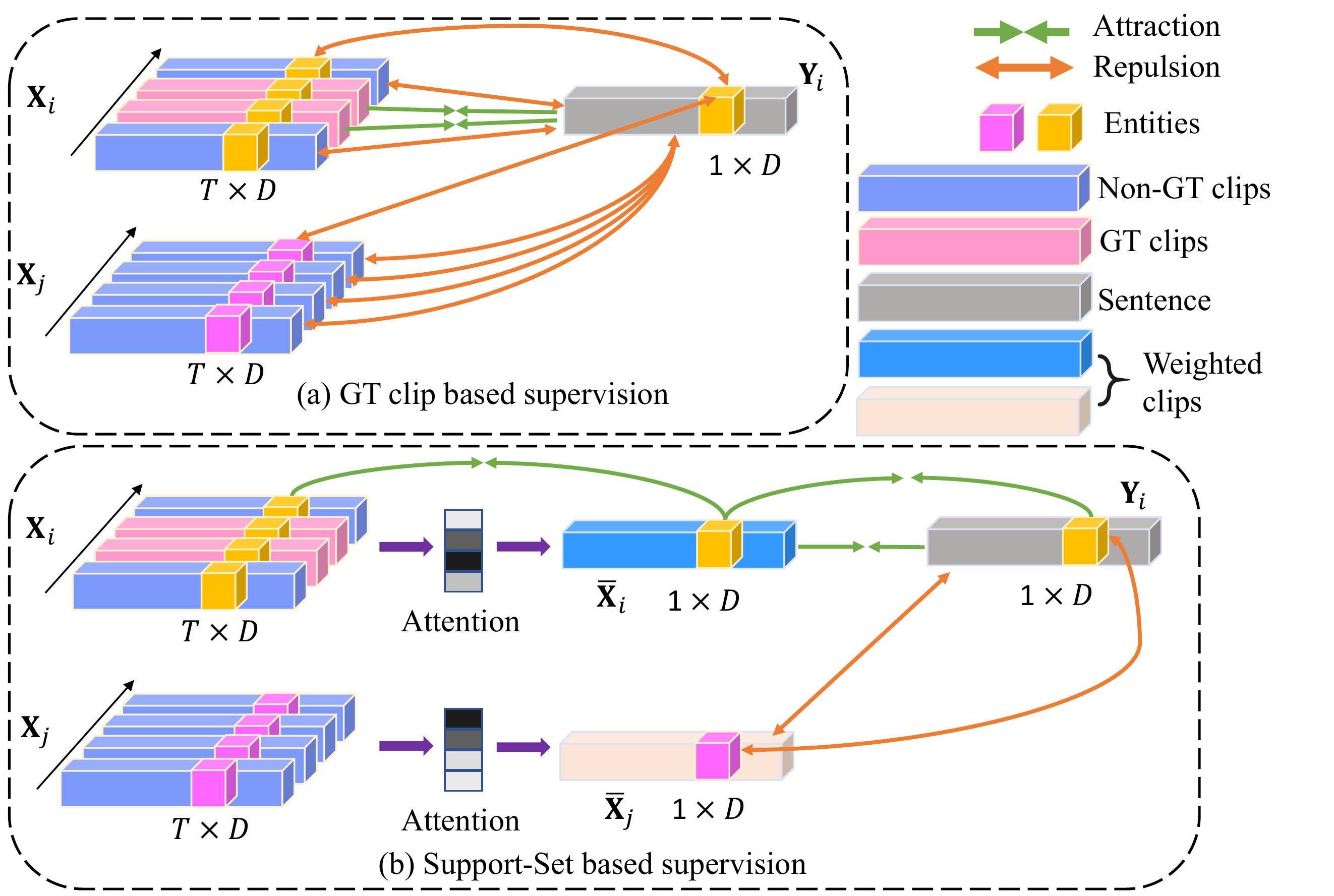}
    \caption{(a) GT clips based supervision. 
    The GT clips based learning aims to encourage the GT clip features to be close with $\mathbf{Y}_i$ and push away the Non-GT clip features. 
    (b) Support-set based supervision. 
    Considering there are also entities from the query in Non-GT clips, \emph{i.e.}, the yellow cube, 
    we maximize the similarity between the weighted feature ($\overline{\mathbf{X}}_i$) and $\mathbf{Y}_i$.
    }
    \label{fig:comparision}
\end{figure}

We name the model training with Eq.~\ref{e:milnce} and Eq.~\ref{e:caption} as the GT clip-based Learning. 
The model will push the sentence feature $\mathbf{Y}_i$ and its corresponding GT clip features to be close, while push $\mathbf{Y}_i$ away from Non-GT clip features. % (BGC feature).
Non-GT clip features of $\mathbf{Y}_i$ contain non-ground-truth clips and clips in other videos.
However, in the same video, the entities may appear in both the GT and Non-GT clips, rather than only GT clips, as shown in Fig.~\ref{fig:comparision} (a).
By simply attracting $\mathbf{Y}_i$ and GT clip features and repulsing $\mathbf{Y}_i$ and Non-GT clip features, the GT clip-based learning would make the same entity (yellow cube in Fig.~\ref{fig:comparision}) in background clips also be far away from that in ground-truth clips.  
Hence, this method is too strict and the learned representations of video clips may be far away even they have similar semantics.

%-------------------------------------------------------------------------
%-----------------------------------------------------------------

\noindent{\bf Support Set-Based Supervision.}
To address the mutual exclusion between the videos and texts as analyzed above, we propose a support-set based supervision approach. 
Our motivation is that different clips in a video may share same semantic entities.
For example, given a sentence query `The person pours some water into the glass' and its corresponding video, the person entity and glass entity appear throughout the video, as shown in Fig.~\ref{fig:mutualexclusion}, and only in GT clips, the action of `pour water' occurs.
Although there are no `pour water' happening in Non-GT clips, the semantics of them are also similar with those of ground-truth ones, \emph{e.g.}, the semantics of `The person pours some water into the glass' is much close to that of `The person hold a glass', rather than that of `Person put a notebook in a bag'. 
If we strictly push away the representations of the Non-GT clips, the model would only extract `pour water' in the video and the text, while ignoring `person' and `glass'.

In order to make the learned representations of Non-GT clips with the same entities also have a certain degree of similarity with the corresponding text, we introduce a {\it support-set}, defined as $S_i$, for each text $L_i$.
The clips in $S_i$ normally have the same entities. In this work, we set all clips in a video as the support-set of its corresponding text, \emph{i.e.}, $S_i = \{ \mathbf{x}_t^i \}_{t=1}^T$, where $\mathbf{x}_t^i \in \mathbb{R}^D$ is the embedding of $v^i_t$.
This is because in video grounding, clips in the same video usually belong to the same scene, and most of the people and things in those clips are similar or even the same.
Based on the support-set $S_i$, we first compute the similarity between all clips in  $S_i$ and $L_i$ and then the clip-wise attention can be obtained as a softmax distribution over clip indices:
\begin{equation}
    a^i_t= \frac{ e^{ \left\langle \mathbf{x}^i_t, \mathbf{Y}_i \right\rangle  / \tau}}{\sum_{\mathbf{x} \in \mathcal{S}_{i}}  e^{\left\langle \mathbf{x}, \mathbf{Y}_i \right\rangle / \tau}  },
    \label{e:attention}
\end{equation}
where, $\left\langle \mathbf{x}^i_t, \mathbf{Y}_i \right\rangle $ is the cosine similarity between $\mathbf{x}^i_t$ and $\mathbf{Y}_i$. 
Then, we compute the weighted average of the embeddings in $S_i$ as follows:
\begin{equation}
   \overline{\mathbf{w}}_i=\sum\limits_{t=1}^T a^i_t \cdot \mathbf{x}_t^i.
    \label{e:weighted}
\end{equation}
After acquiring $\overline{\mathbf{w}}_i$, we can redefine the positive samples and the negative samples in $\mathcal{B}$.
Concretely, we set the $\{(\overline{\mathbf{w}}_i, \mathbf{Y}_i)\}_{i=1}^B$ as positive samples, and other pairs in the batch as negative ones, i.e, $\overline{\mathcal{N}}_i=\{(\overline{\mathbf{w}}_i, \mathbf{Y}_j)\}_{i \neq j}$.
Then, the contrastive objective can be defined as follows:
\begin{equation}
     \mathcal{L}^{\text{contrast}}=-\sum\limits_{i=1}^{B} \log \left(\frac{ e^{ \overline{\mathbf{w}}_i^{\top} \mathbf{Y}_i / \tau} }{e^{ \overline{\mathbf{w}}_i^{\top} \mathbf{Y}_i / \tau} +\sum\limits_{\left(\mathbf{x}^{\prime}, \mathbf{y}^{\prime}\right) \in \overline{\mathcal{N}}_{i}} e^{ {\mathbf{x}^{\prime}}^{\top} {\mathbf{y}^{\prime}} / \tau}  }\right),
     \label{e:ssnce}
\end{equation}
and the caption objective is:

\begin{equation}
   \mathcal{L}^{\text{caption}} = -\frac{1}{B} \sum_{i=1}^{B} \log p\left(l_{i} \mid \overline{\mathbf{w}}_{i}\right).
    \label{e:sscaption}
\end{equation}

We name the model training with Eq.~\ref{e:ssnce} and Eq.~\ref{e:sscaption} as the support-set based supervision.
As Fig.~\ref{fig:comparision} (b) shows, besides pushing the sentence feature $\mathbf{Y}_i$ and its corresponding GT clip feature to be close, the representations of the same entity (yellow cube) in both Non-GT clip features and the sentence feature are also be attracted.

\noindent{\bf Comparison between \cite{patrick2020support} and Ours}. The main differences with SS are two-fold:
i) Motivations.
Our goal is to apply the cross-supervision to capture the relations between the visual semantics and textual concepts.
While \cite{patrick2020support} aims to improve video-text representations by relaxing the contrastive objective;
ii) Solutions.
In SS, the cross-captioning objective is to relax the strict contrastive objective, hence it's an adversarial relationship actually.
While in Scsc, our two objectives are in a cooperative relationship because they both aim to learn the video-text relations.
Furthermore, our contrastive objective is build on global video features encoded by support-set, while \cite{patrick2020support} applies a triplet ranking loss based on local clip features. 

\subsection{Several Kinds of Support-Set}
\label{sec:differentss}
The support-set based supervision contains two basic operations: 
(a) {\it the construction of the support-set} $\mathcal{S}_i$;  
and (b) {\it the function to map the support-set to a weighted embedding} $\overline{\mathbf{w}}_i$.
In this section, we explore three kinds of functions to construct the support-set: 
{\bf (a) video-level support set (V-SS)}: 
we set all clips in a video as the support, \emph{i.e.}, $\mathcal{S}_i = \{ \mathbf{x}_t^i \}_{t=1}^T$;
{\bf (b) ground-truth-level support set (GT-SS)}, which only contains the GT clips, \emph{i.e.}, $\mathcal{S}_i = \mathcal{M}_i$;
{\bf (c) Non-GT level support set (Non-GT-SS)}: which only contains the Non-GT clips, \emph{i.e.}, $\mathcal{S}_i = \overline{\mathcal{M}}_i$.

By these functions, we compare six ways as follows:
{\bf (a) cross-attention (CA).} The function is computed by Eq.~\ref{e:attention} and Eq.~\ref{e:weighted}; 
{\bf (b) self-attention (SA).} We first concatenate the clips in $\mathcal{S}_i$ along clip indices to obtain $\mathbf{S}_i$, then we compute the similarity matrix of $\mathbf{S}_i$ by $\mathbf{Q}_i = \mathbf{S}_i^{\top} \mathbf{S}_i / \tau$. 
The $t$-th vector of $\mathbf{Q}_i$ is $\mathbf{q}^i_t \in \mathbf{R}^D$. 
Sum all the elements of $\mathbf{q}^i_t$ to obtain the summed scalar $z^i_t$. Then we obtain the clip-wise attention as follows:

\begin{equation}
    a^i_t= \frac{e^{z^i_t}}{\sum_{z \in \mathcal{Z}_i} e^{z}},
\end{equation}

where $\mathcal{Z}_i$ is the set of all $z^i_t$ for $\mathbf{Q}_i$. Finally, $\overline{\mathbf{w}}_i$ can be obtained by Eq.~\ref{e:weighted}.
{\bf (c) fully-connected layer (FC).} In this way, after concatenating the clips in $\mathcal{S}_i$ along clip indices, the concatenated feature $\mathbf{S}_i$ is converted to $\overline{\mathbf{w}}_i$ by a fully-connected layer.
{\bf (d) Convolutional layer (Conv).} Similar to FC, we fed $\mathbf{S}_i$ into a convolutional layer to acquire $\overline{\mathbf{w}}_i$.
{\bf (e) Max-pooling (MP).} In this way, after concatenating the clips in $\mathcal{S}_i$ along clip indices, the concatenated feature $\mathbf{S}_i$ is fed into a max-pooling layer to acquire $\overline{\mathbf{w}}_i$.
{\bf (f) Average-pooling (AP).} Similar to MP, we fed $\mathbf{S}_i$ into a average-pooling layer to acquire $\overline{\mathbf{w}}_i$.
%-------------------------------------------------------------------------
%--------------------------------------------
\begin{table}[t]
\centering
\caption{Ablation study of different supervision methods on the Charades-STA dataset.}\smallskip
\label{table:methods}
\resizebox{1.0\columnwidth}{!}{
\begin{tabular}{ c   | c  c | c | c |c |c}
\hline 
\hline
\multirow{2}{*}{Model }   &\multirow{2}{*}{ $\mathcal{L}^{\text{contrast}}$ }  & \multirow{2}{*}{$\mathcal{L}^{\text{caption}}$ } & \multicolumn{2}{c|}{$Rank 1@$} &\multicolumn{2}{c}{$Rank 5@$}\\
 \cline{4-7}
  &  & & 0.5& 0.7& 0.5 & 0.7\\
\hline
2D-TAN \cite{zhang2019learning} &
 & & 
50.62 &	 28.71 & 79.92&	48.52 \\
\hline
\multirow{3}{*}{2D-TAN+GTC}&
\checkmark & &
 54.77	& 31.63	& 86.28 &	55.07 \\
\cline{2-7}
 &
 & \checkmark &
51.72	& 29.35&	83.66&	52.12\\
 \cline{2-7}
 &
  \checkmark  & \checkmark &
  55.40	& 32.15	& 87.07 &55.62\\
 \hline
\multirow{3}{*}{2D-TAN+SS}&
\checkmark & &
  56.19	 &32.03 & 87.95 & 56.05  \\
\cline{2-7}
 &
 & \checkmark &
53.12&	30.05	&85.19	& 53.28 \\
 \cline{2-7}
 &
  \checkmark  & \checkmark &
{\bf 56.97}&{\bf 32.74} &{\bf 88.65}&	{\bf 56.91}\\
  \hline
\hline

LGI \cite{mun2020local}&
& &
59.46 & 35.48 & - & -  \\
\hline
\multirow{3}{*}{LGI+GTC}&
\checkmark & &
59.63	& 35.71 &- &- \\
\cline{2-7}
 &
 & \checkmark &
 59.88 & 35.92 & - & -\\
 \cline{2-7}
 &
  \checkmark  & \checkmark &
 60.02 & 36.11 & - & -\\
 \hline
\multirow{3}{*}{LGI+SS}&
\checkmark & &
60.09&36.32&-&- \\
\cline{2-7}
 &
 & \checkmark &
  60.53&36.75&-&- \\
 \cline{2-7}
 &
  \checkmark  & \checkmark &
  {\bf 60.75}	&  {\bf 37.29} & - &- \\
  \hline

\hline
\end{tabular}
}
\end{table} 
%--------------------------------------------
%----------------------------------------------------------------------------
\section{Experiments}
\label{sec:experiments}
%-----------------------------------------------------------------------------
\subsection{Datasets}
\label{sec:datasets}
\noindent{\bf TACoS.} TACoS is collected by Regneri {\it et al.} \cite{regneri2013grounding} which consists of 127 videos on cooking activities with an average length of $4.79$ minutes for video grounding and dense video captioning tasks. 
We follow the same split of the dataset as Gao {\it et al.} \cite{gao2017tall} for fair comparisons.

\noindent{\bf Charades-STA.} Charades is originally collected for daily indoor activity recognition and localization \cite{sigurdsson2016hollywood}, which consists of $9,848$ videos.
Gao {\it et al.} \cite{gao2017tall} build the Charades-STA by annotating
the temporal boundaries and sentence descriptions of Charades\cite{sigurdsson2016hollywood}.

\noindent{\bf ActivityNet-Captions.} ActivityNet \cite{caba2015activitynet} is a large-scale dataset which is collected for video recognition and temporal action localization \cite{lin2018bsn,chao2018rethinking,gao2017cascaded,ding2020weakly,shou2016temporal,narayan20193c,paul2018w}.
Krishna {\it et al.} \cite{krishna2017dense} extend ActivityNet to ActivityNet-Captions for the dense video captioning task. 
%--------------------------------------------
\begin{table*}[t]
\centering
\caption{Ablation study of different kinds of construction methods and function methods on the Charades-STA dataset.}\smallskip
\label{table:kinds}
\resizebox{1.5\columnwidth}{!}{
\begin{tabular}{ c  | c  c  c c  c  c | c | c |c |c}
\hline 
\hline
\multirow{2}{*}{Construction method } & \multicolumn{6}{c|}{Function method} & \multicolumn{2}{c|}{$Rank 1@$} &\multicolumn{2}{c}{$Rank 5@$}\\
 \cline{2-11}

  & CA & SA & FC & Conv & MP & AP& 0.5& 0.7& 0.5 & 0.7\\
\hline
 \multirow{6}{*}{V-SS } &
\checkmark &             &          &  &   &  & 
{\bf 56.97} &  {\bf 32.74} & {\bf 88.65 } & {\bf56.91 } \\
\cline{2-11}
 &
 &       \checkmark      &          &  &   &  &  
54.88 &	30.98 &	86.56 &	54.92            \\
\cline{2-11}
  &
 &           &      \checkmark      &  &   &  &   
54.91	& 31.25 &	86.75	& 55.01            \\
\cline{2-11}
  &
 &           &       &      \checkmark &   &  &   
54.89 &	31.08	& 86.75	& 54.73            \\
\cline{2-11}
  &
 &           &       &      &   \checkmark &  &   
53.35 &	30.64 &	86.13 &	54.35              \\
\cline{2-11}
  &
 &           &       &      &    & \checkmark  &   
53.14 &	30.36	& 86.10	& 54.13 \\
\hline
\hline
\multirow{6}{*}{GT-SS } &
\checkmark &             &          &  &   &  & 
55.91 &	32.03 &	88.12	& 55.25           \\
\cline{2-11}
 &
 &      \checkmark       &          &  &   &  & 
54.89 &	31.23 &	87.11	& 54.40                 \\
\cline{2-11}
 &
 &             &     \checkmark     &  &   &  & 
54.90 &	31.17	& 87.10	& 54.85         \\
\cline{2-11}
 &      
 &             &          &  \checkmark &   &  & 
54.85 &	 31.10 &	87.52	& 54.88             \\
\cline{2-11}
 &
 &             &          &   & \checkmark  &  & 
53.62 &	30.80	& 86.54 &	54.79          \\
\cline{2-11}
 &
 &             &          &   &   & \checkmark &  
53.70 &	30.91 &	86.78	& 54.88         \\
\hline
\hline
\multirow{6}{*}{Non-GT-SS }&
\checkmark &             &          &   &   &  &  
 50.12 &	28.96	& 85.82 &	52.78    \\    
\cline{2-11}
 &
 &         \checkmark    &          &   &   &  &  
48.55 &	26.64	& 83.27 &	50.62    \\    
\cline{2-11}
 &
 &             &      \checkmark    &   &   &  &  
48.52 &	26.56	& 83.31 &	50.64    \\    
\cline{2-11}
 &
 &             &         &   \checkmark  &   &  &  
  48.29 &	26.44 &	81.13	&  50.44  \\
\cline{2-11}
 &
 &             &         &    &  \checkmark  &  &  
48.87 &	26.57	& 83.40 &	50.60    \\    
\cline{2-11}
 &
 &             &         &    &    & \checkmark &  
48.33 &	26.48	& 83.34 &	50.52    \\    
\hline
\end{tabular}
}
\end{table*} 
%--------------------------------------------
%-------------------------------------------------------------------------
\begin{figure}
    \centering
    \includegraphics[width=1\columnwidth,height=0.17\textheight]{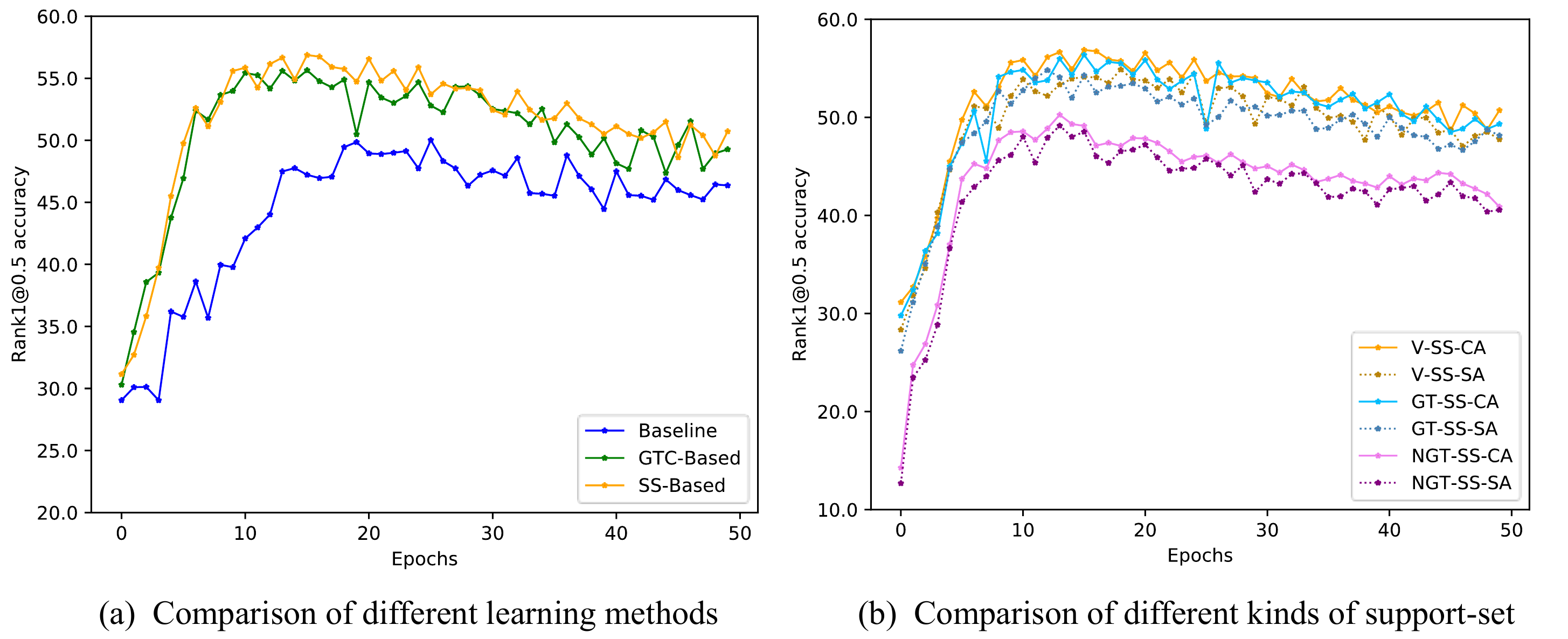}
    \caption{(a) Comparison of the accuracy curve of different learning methods. (b) Comparison of the accuracy curve of different kinds of support-set.
    }
    \label{fig:accuracy}
\end{figure}
%-----------------------------------------------------------------
%-----------------------------------------------------------------------------
\subsection{Implementation details}
\label{sec:implementation}

\noindent{{\bf Evaluation metric.}} 
For fair comparisons, we follow the setting as previous work \cite{gao2017tall} and evaluate our model by computing $Rank~n@m$.
Specifically, it is defined as the percentage of queries having at least one correct grounding prediction in the top-$n$ predictions, and the grounding prediction is correct when its IoU with the ground truth is larger than $m$. 
Similar to \cite{zhang2019learning}, we evaluate our method with specific settings of $n$ and $m$ for different datasets. 

\noindent{{\bf Feature Extractor.}} 
For a fair comparison, we extract video features following previous works \cite{zhang2019learning,zeng2020dense}. 
Specifically, We use the C3D \cite{tran2015learning} network pre-trained on Sports-1M \cite{karpathy2014large} as the feature extractor. 
For Charades-STA, we also use VGG \cite{simonyan2014very}, C3D \cite{tran2015learning} and I3D \cite{carreira2017quo} feature  to compare out results with \cite{gao2017tall,zhang2019learning}.
We divided the video into segments and each contains fixed number frames. 
The input of C3D network is a segment with $16$ frames for three datasets.
When using VGG feature for Charades-STA, the number of frames in a segment is set to $4$. Non maximum suppression (NMS) with a threshold of $0.5$ is applied during the inference. $\tau$ is set to 0.1. For Charades-STA, $\lambda_1$ and $\lambda_2$ are set to 0.1, and for TACoS, $\lambda_1$ and $\lambda_2$ are set to 0.001.

\noindent{{\bf Baseline Model.}} 
Our work is built on two current state-of-the-art models for video grounding, 2D temporal adjacent network (2D-TAN) \cite{zhang2019learning} and local-global video-text interactions (LGI) \cite{mun2020local}.
%-------------------------------------------------------------------------
\begin{figure}
    \centering
    \includegraphics[width=1\columnwidth,height=0.15\textheight]{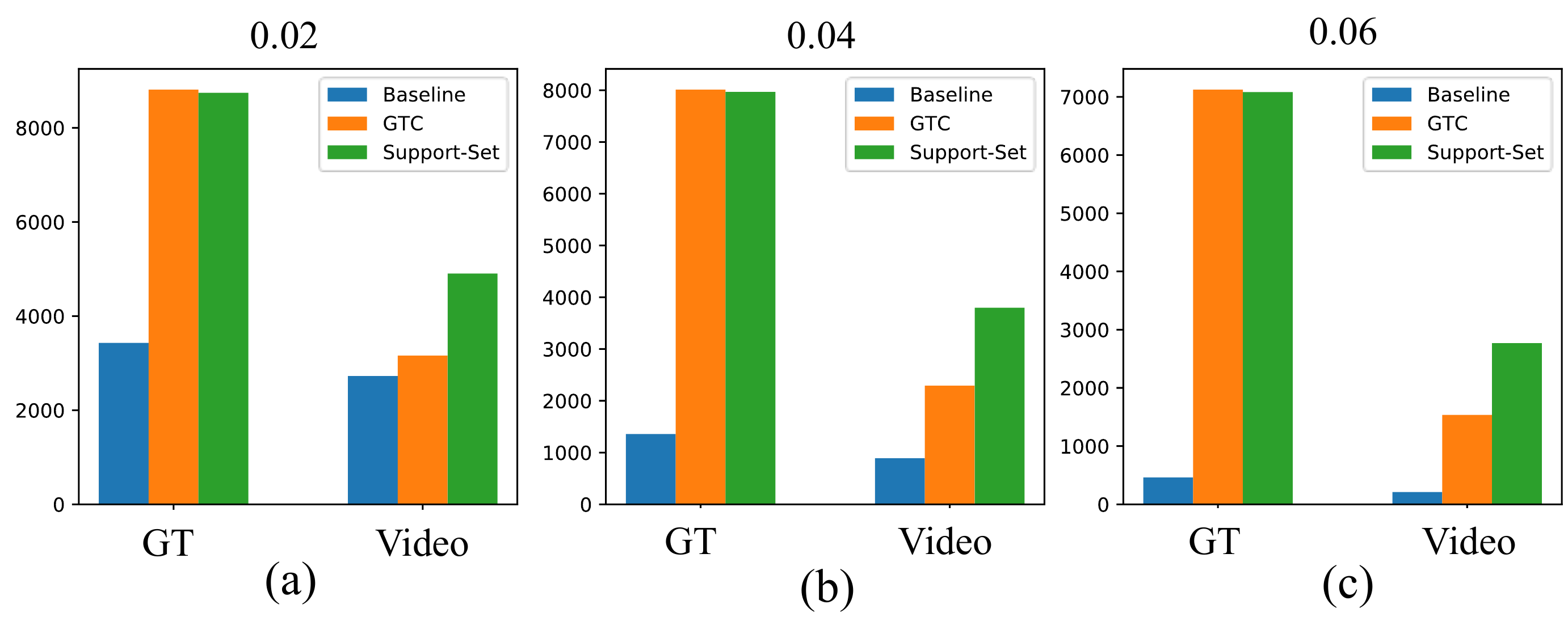}
    \caption{Comparison of recalls of high-related video-text pairs under different thresholds.
    }
    \label{fig:recall}
\end{figure}
%-----------------------------------------------------------------

\noindent{{\bf Training settings.}}
We use Adam \cite{kingma2014adam} with learning rate of $1.6 \times 10^{-2}$ and batch size of $256$ for optimization. 
We decay the learning rate with ReduceLROnPlateau function in Pytorch \cite{32}. All of our models are implemented by PyTorch  and trained under the environment of Python 3.6 on Ubuntu 16.04.
%------------------------------------------------------------------
\begin{table}[t]
\centering
\caption{Comparisons with state-of-the-arts on Charades-STA.}\smallskip
\label{table:charades-sta}
\resizebox{0.9\columnwidth}{!}{
\begin{tabular}{c|c|c |c |c | c }
\hline
\hline
    \multirow{2}{*}{Methods} & {\multirow{2}{*}{Feature}}& \multicolumn{2}{c|}{$Rank 1@$} &\multicolumn{2}{c}{$Rank 5@$}  \\
    \cline{3-6}
 & & 0.5 & 0.7 &0.5  & 0.7 \\
   \hline
VAL \cite{song2018val}& VGG & 23.12 & 9.16 & 61.26 & 27.98 \\
ACL-K \cite{ge2019mac}& VGG & 30.48 & 12.20 & 64.84 & 35.13 \\
TripNet \cite{hahn2019tripping}& VGG & 36.61 & 14.50 & - & - \\ 
DRN \cite{zeng2020dense}& VGG & 42.90 &23.68 &{\bf 87.80} & {\bf 54.87}\\
\hline
2D-TAN \cite{zhang2019learning}& VGG & 39.70 & 23.31 & 80.32 & 51.26 \\
2D-TAN + ours& VGG & {\bf 43.15} & { \bf 25.54} &  84.26 &  54.17 \\
\hline
LGI \cite{mun2020local} & VGG & 41.72 & 21.48 &- &- \\
LGI + ours& VGG &  43.68 & 23.22 & - & - \\

 \hline
MAN \cite{2018MAN} & I3D & 46.53 & 22.72& 86.23 & 53.72 \\
DRN \cite{zeng2020dense}& I3D & 53.09 & 31.75 & {\bf 89.06} & {\bf 60.05}\\
\hline
2D-TAN \cite{zhang2019learning} & I3D & 50.62 &  28.71  & 79.92 & 48.52 \\
2D-TAN + Ours  &I3D &  56.97 &  32.74 &  88.65 &  56.91 \\
\hline
LGI \cite{mun2020local}& I3D & 59.46 & 35.48 & - & -\\
LGI + Ours& I3D & {\bf 60.75} & {\bf36.19} & - & - \\
 
\hline
\end{tabular}}
\end{table} 
%------------------------------------------------------------------
%-----------------------------------------------------------------------------
\subsection{Ablation Study}
\label{sec:ablation}
In this section, all presented results are on Charades-STA \cite{gao2017tall} with I3D \cite{carreira2017quo} features.
For convenience, we use `GTC' and `SS' to refer to GT clip and support-set based supervision in following experiments.
%---------------------------------------------------------------------------
\begin{table*}[t]
\centering
\caption{Comparisons with state-of-the-arts on TACoS and ActivityNet-Captions.}\smallskip
\label{table:tacosandanet}
\resizebox{1.8\columnwidth}{!}{
\begin{tabular}{c|c |c |c | c| c| c | c |c |c | c| c| c}
\hline
\hline
&\multicolumn{6}{c|}{TACoS}&\multicolumn{6}{c}{ActivtiyNet-Captions}\\
\hline
    \multirow{2}{*}{Methods}& \multicolumn{3}{c|}{$Rank 1@$} &\multicolumn{3}{c|}{$Rank 5@$} & \multicolumn{3}{c|}{$Rank 1@$} &\multicolumn{3}{c}{$Rank 5@$} \\
    \cline{2-13}
  & 0.1 & 0.3 &0.5  & 0.1 & 0.3 &0.5& 0.3 & 0.5 &0.7  & 0.3 & 0.5 &0.7\\
   \hline
TGN \cite{chen2018temporally} &41.87 & 21.77 & 18.9 & 53.40 & 39.06 & 31.02&43.81 & 27.93 & - & 4.56 & 44.20   \\
ACRN \cite{liu2018attentive} & 24.22 & 19.52 & 14.62 & 47.42 & 34.97 & 24.88& 49.70 & 31.67 & 11.25 & 76.50 & 60.34 & 38.57 \\
CMIN \cite{zhang2019cross} & 32.48 & 24.64 & 18.05 & 62.13 & 38.46 & 27.02 & - & - & -&- &-&-\\
QSPN \cite{xu2019multilevel} & 25.31 & 20.15 & 15.23 & 53.21 & 36.72 & 25.30& 52.13 & 33.26 & 13.43 & 77.72 & 62.39 & 40.78 \\ 
ABLR \cite{yuan2019find} & 34.70 & 19.50 & 9.40 & - & - & -& 55.67 & 36.79 & - & -&-&-  \\ 
DRN \cite{zeng2020dense} & -& -&23.17 & - & - & 33.36& - & 45.45 & 24.36 & - & 77.97 & 50.30 \\
HVTG \cite{chenhierarchical}& - & - & -&- &-&-& 57.60 & 40.15 & 18.27 &- &-&-\\
\hline
2D-TAN \cite{hahn2019tripping} & 47.59 & 37.29 & 25.32 & 70.31 & 57.81 & 45.04& 59.45 & 44.51 & 26.54 & 85.53 & 77.13 & 61.96 \\
2D-TAN + Ours & {\bf 50.78 } & {\bf 41.33} & {\bf 29.56} & {\bf 72.53} & {\bf 60.65} & {\bf 48.01} & {\bf 61.35} & {\bf 46.67} & {\bf 27.56} & {\bf 86.89} & {\bf 78.37} & {\bf  63.78} \\
\hline
LGI \cite{mun2020local}  & - & - & - & - & - & - & 58.52 & 41.51 & 23.07 & - & - & - \\

LGI + Ours  & - & - & - & - & - & - & 59.75 & 43.62 & 25.52 & - & - & - \\
 \hline
\end{tabular}}
\end{table*} 
%---------------------------------------------------------------------

%-----------------------------------------------------
\noindent{\bf Comparison of different supervision methods.} 
In this ablation study (Table \ref{table:methods}), we compare different learning methods, proposed in Section \ref{sec:supportset}, including GT clip based supervision and support-set based supervision.
It is clear from Table \ref{table:methods}, the performance of SS outperforms that of GTC with large margins. 
What's more, the contrastive objective $\mathcal{L}^{\text{contrast}}$ bring larger  performance improvement than the caption one $\mathcal{L}^{\text{caption}}$.
Combining both the contrastive objective and the caption objective, our model obtain the best performance.
The interactions of videos and text in 2D-TAN \cite{zhang2019learning} is Hadamard product, and those in LGI \cite{mun2020local} is in a coarse-to-fine manner which is more fine-grained than 2D-TAN.
For 2D-TAN, the interaction and grounding modules are compute the similarity between video clips and text to predict target intervals, which is very similar to the objective of cross-supervision.  Hence, our method achieves larger improvement in 2D-TAN.
As Fig.~\ref{fig:accuracy} (a) indicates, with an extra Cscs branch, besides the higher performance, the model converges faster than the baseline method.

%---------------------------------------------------------------
\noindent{\bf Comparison with different kinds of the support-set.} 
In this ablation study, we compare different kinds of construction methods and function methods of the support-set. Table \ref{table:kinds} presents the performance of different kinds of the support-set on the Charades-STA dataset. Specifically, we compare three types of construction methods: (a) V-SS, (b) GT-SS, (c) Non-GT-SS and six ways of function methods: (a) CA, (b) SA, (c) FC, (d) Conv, (e) MP, (e) AP (See details in Section \ref{sec:differentss} ). Our proposed method (V-SS + CA) achieve the best performance. The V-SS way can make the learned representation explore more similar entities in non-ground-truth clips. CA aims to find the high similarity between videos and text, while other function methods (\emph{e.g.}, SA, FC, etc.) only considering the single modality information (i.e, videos). Hence, CA is more effective in the support-set. Since Non-GT-SS only contains non-ground-truth clips, the learned representation of the videos and text would let the ground-truth clips have dissimilar semantics, resulting in poor performance in video grounding. The comparison of accuracy curves presented in Fig.~\ref{fig:accuracy} (b).
%
%--------------------------------------------------------

\noindent{\bf Recalls of high-related video-text pairs.}
In order to verify that the proposed approach can enhance the correlation between text and videos, we present the recalls of high-similar video-text pairs under different thresholds ($0.02$, $0.04$ and $0.06$) in Fig.~\ref{fig:recall}. `Video' indicates the average similarity between clips in the whole video and text, and `GT' is the average similarity between GT clips and text.
It is clear that, adding cross-supervision module can significantly improve the similarity between the video and text.
Support-set based approach can have a more generalized representation, compared with the GT clip based learning.

%-----------------------------------------------------------------------------
\subsection{Comparison with the State-of-the-Arts}
\label{sec:comparison}
%-----------------------------------------------------------------------------
We conduct experiments on TACoS, Charades-STA and ActivityNet-Captions datasets to compare with several State-Of-the-Art (SOTA) approaches. From Table \ref{table:charades-sta} and Table \ref{table:tacosandanet}, it clearly shows that the proposed method can largely improve the SOTA models, \emph{i.e.}, 2D-TAN \cite{zhang2019learning} and LGI \cite{mun2020local}, almost without any extra inference cost. 
We can also see that Sscs achieve smaller gain with LGI. The reasons may be that
LGI is a \emph{regression} based method that directly regress the boundaries, while 2D-TAN is a \emph{comparison and selection} based method that compares text with dense proposals and selects the best one.
In Scsc, SS is built on a contrastive objective, which has a similar spirit with 2D-TAN, hence it achieves larger gains on 2D-TAN. 
Furthermore, with 2D-TAN, SS obtains larger gains by $6.35\%$ and $4.24\%$ on Charades and TACoS than that $2.16\%$ on Activities.
We think it because that Charades and TACoS have static and smooth backgrounds and simple actions, while ActivityNet is more complex and diverse.
Thus the improvement on ActivityNet is relatively small.

%----------------------------------------------------------------------------
%-------------------------------------------------------------------------
\begin{figure}
    \centering
    \includegraphics[width=1\columnwidth,height=0.18\textheight]{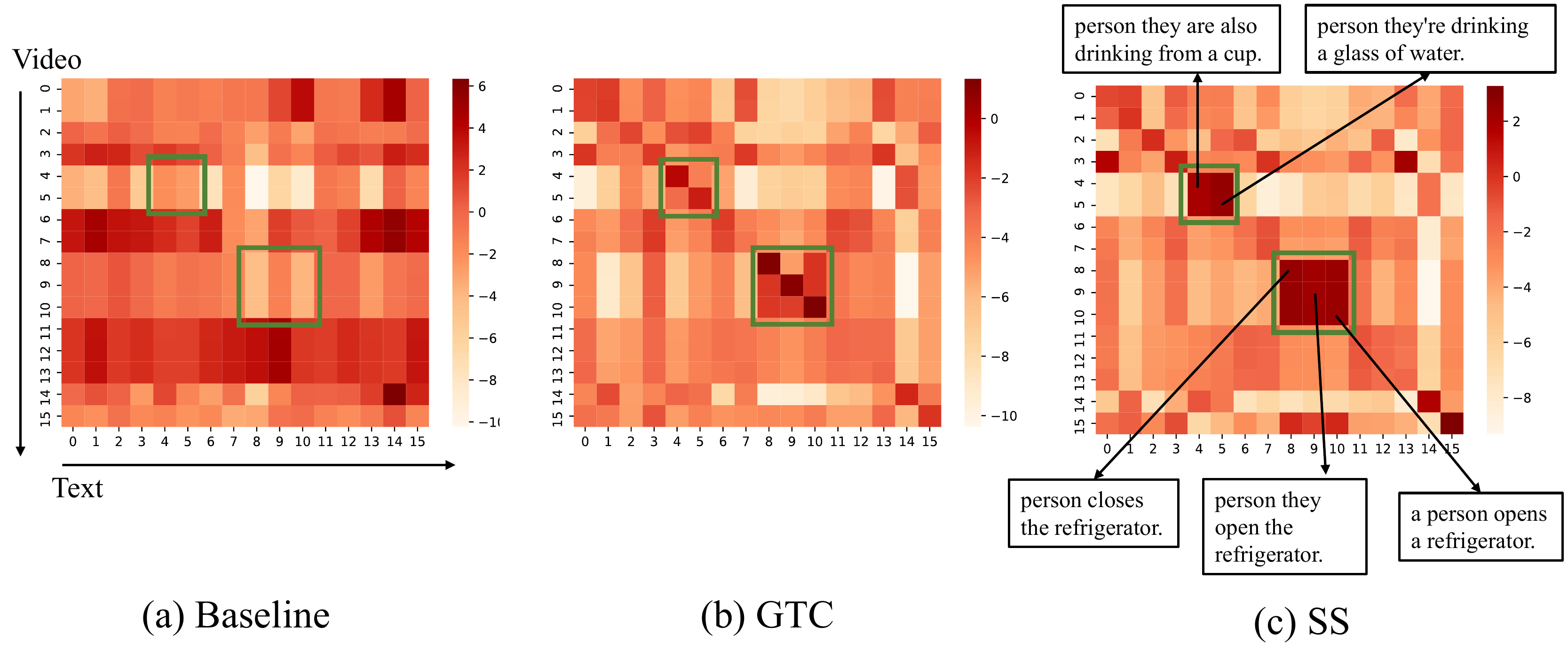}
    \caption{similarity matrix of (a) the baseline, (b) GTC and (c) SS. We present 16 video-text sample pairs.}
    \label{fig:sim}
\end{figure}
%-----------------------------------------------------------------
\subsection{Qualitative analysis}
In this section, we present some qualitative results on Charades-STA.
We present the similarity matrix of video-text pairs in Fig.~\ref{fig:sim}. It is clear that the baseline model can not capture the semantic similarity of video-text pairs even they come from the same sample (see Fig.~\ref{fig:sim} (a) ). On the contrary, the similarity score of videos and text from the same sample would be higher than others. Compared with GTC, SS can also capture the related semantics pairs, even they are not from the same sample. As Fig.~\ref{fig:sim} shows, the text in $4$-th and $5$-th samples have similar semantics, and the similarity of the corresponding videos are also high, which are not found in the baseline model and GTC.

\begin{figure}
    \centering
    \includegraphics[width=1\columnwidth,height=0.15\textheight]{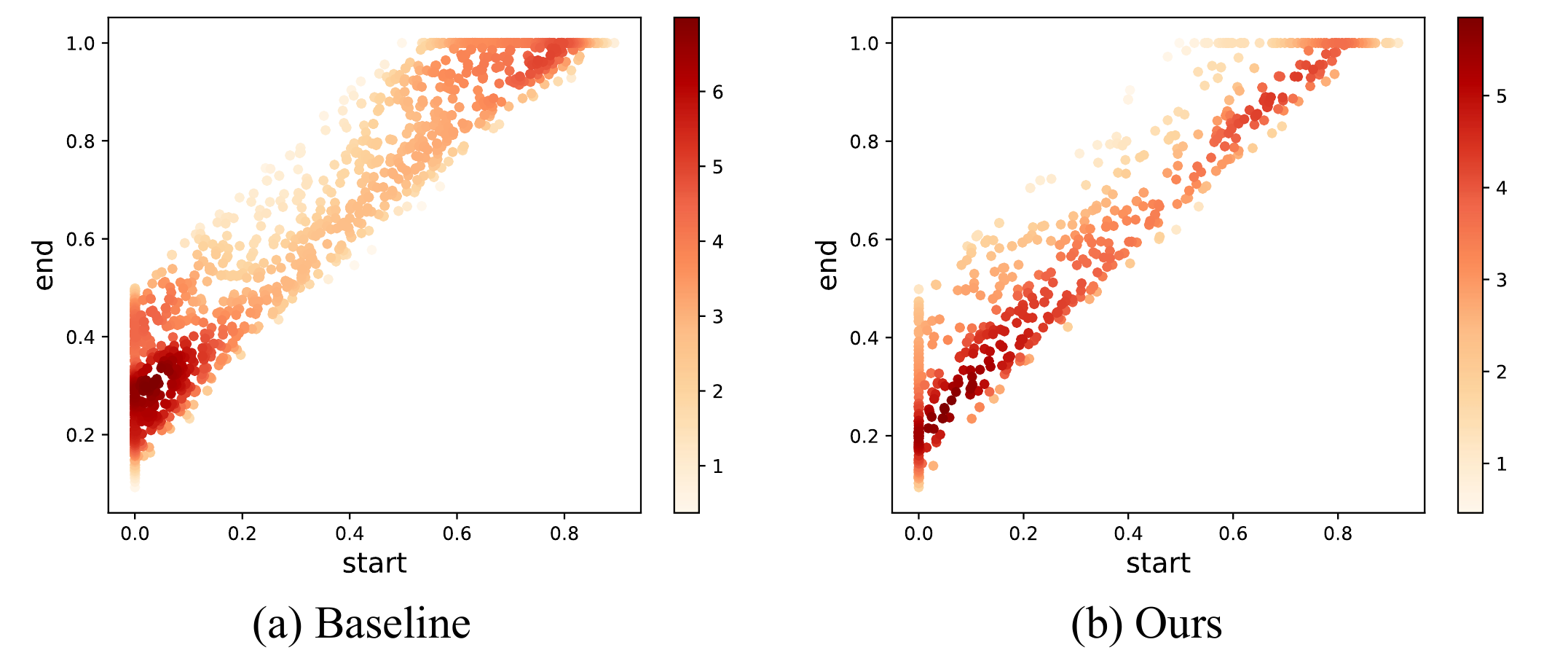}
    \caption{
        (a) The distributions of successfully predicted time intervals by the baseline. (b) The distributions of our model additionally success-predicted time intervals, compared with the baseline model.
    }
    \label{fig:distribution}
\end{figure}

Fig.~\ref{fig:distribution} shows the successfully predicted time interval distributions. It is clear that most of the baseline model predicted time intervals are generally concentrated at the beginning of the video, and the duration of the fragments are about $20\%-40\%$ of the total length of the video, as shown in Fig.~\ref{fig:distribution} (a). Compared with the time intervals predicted by the baseline model, the proposed method can find more time intervals occurring in the middle of the videos, also the duration of the time intervals are shorter, which is indicated in Fig.~\ref{fig:distribution} (b). This reason is that the proposed method can learned better video-text representations, thus we can find more time intervals that are difficult to locate.

%-----------------------------------------------------------------
%----------------------------------------------------------------------------
\section{Conclusion}
In this paper, we introduce a Support-Set Cross-Supervision (Sscs) Module as an extra branch to video grounding to extract the correlation between videos and text.
By conducting the contrastive and caption objective to the clip-level and sentence features in a shared space, the learned two-modality features are enforced to become similar, only if the semantics of them are related. 
To address the mutual exclusion of entities, we improve the cross-supervision with the support-set to collect all important visual clues from the whole video.
The experimental results shows the proposed method can greatly improve the performance of the state-of-the-art backbones almost without any extra inference cost, and the ablation study verify the effective of the support-set.

%--------------------------------------------------------------------------
\noindent\paragraph{Acknowledgements.}
This work was supported in part by the National Key Research and Development Program of China under Grant 2018AAA0103202; in part by the National Natural Science Foundation of China under Grant Grants 62036007, 61922066, 61876142, 61772402, and 6205017; in part by the Fundamental Research Funds for the Central Universities.
{\small
\bibliographystyle{ieee_fullname}
\bibliography{egbib}
}

\end{document}